\documentclass[11pt]{article}

\usepackage{acl}

\usepackage{times}
\usepackage{latexsym}

\usepackage[T1]{fontenc}

\usepackage[utf8]{inputenc}

\usepackage{microtype}

\usepackage{inconsolata}

\usepackage{graphicx}

\usepackage{xspace}
\usepackage{adjustbox}
\usepackage{pgfplots}  
\usepackage{tcolorbox}
\usepackage{minted}
\usepackage{amsmath}
\usepackage{listings}
\usepackage{booktabs}
\usepackage{xcolor}
\usepackage{bm}
\usepackage{enumitem}
\usepackage{multirow}
\usepackage{amssymb}

\usepackage{color}
\definecolor{keywordcolor}{rgb}{0.7, 0.1, 0.1}   
\definecolor{tacticcolor}{rgb}{0.0, 0.1, 0.6}    
\definecolor{commentcolor}{rgb}{0.4, 0.4, 0.4}   
\definecolor{symbolcolor}{rgb}{0.0, 0.1, 0.6}    
\definecolor{sortcolor}{rgb}{0.1, 0.5, 0.1}      
\definecolor{attributecolor}{rgb}{0.7, 0.1, 0.1} 
\definecolor{lightgray}{rgb}{0.95, 0.95, 0.95}

\usepackage[utf8]{inputenc}
\usepackage{newunicodechar}
\newunicodechar{₀}{\ensuremath{_0}}
\newunicodechar{₁}{\ensuremath{_1}}
\newunicodechar{₂}{\ensuremath{_2}}
\newunicodechar{₃}{\ensuremath{_3}}
\newunicodechar{₄}{\ensuremath{_4}}
\newunicodechar{₅}{\ensuremath{_5}}
\newunicodechar{₆}{\ensuremath{_6}}
\newunicodechar{₇}{\ensuremath{_7}}
\newunicodechar{₈}{\ensuremath{_8}}
\newunicodechar{₉}{\ensuremath{_9}}
\newunicodechar{ℚ}{\ensuremath{\mathbb{Q}}}
\newunicodechar{≠}{\ensuremath{\neq}}
\newunicodechar{ℤ}{\ensuremath{\mathbb{Z}}}
\newunicodechar{✝}{\ensuremath{\dagger}}
\newunicodechar{∣}{\ensuremath{\mid}}
\newunicodechar{ℝ}{\ensuremath{\mathbb{R}}}
\newunicodechar{∧}{\ensuremath{\land}}
\newunicodechar{≤}{\ensuremath{\leq}}
\newunicodechar{⊢}{\ensuremath{\vdash}}
\newunicodechar{ℂ}{\ensuremath{\mathbb{C}}}
\newunicodechar{｜}{|}
\newunicodechar{≥}{\ensuremath{\geq}}
\newunicodechar{λ}{\ensuremath{\lambda}}
\newunicodechar{∞}{\ensuremath{\infty}}
\newunicodechar{∫}{\ensuremath{\int}}
\newunicodechar{ℕ}{\ensuremath{\mathbb{N}}}
\newunicodechar{∀}{\ensuremath{\forall}}
\newunicodechar{∃}{\ensuremath{\exists}}
\newunicodechar{∈}{\ensuremath{\in}}
\newunicodechar{↔}{\ensuremath{\leftrightarrow}}
\newunicodechar{∑}{\ensuremath{\sum}}
\newunicodechar{∨}{\ensuremath{\lor}}
\newunicodechar{→}{\ensuremath{\rightarrow}}
\newunicodechar{·}{\ensuremath{\cdot}}
\newunicodechar{σ}{\ensuremath{\sigma}}

\newcommand{\method}{\textbf{{Diffusion-Proof}}\xspace}

\title{Diffusion-Proof: Recipe for Formal Theorem Proving Beyond Auto-Regressive Generation}

\author{
 \textbf{Ruida Wang\textsuperscript{1}},
 \textbf{Rui Pan\textsuperscript{1}},
 \textbf{Pengcheng Wang\textsuperscript{1}}, 
 \textbf{Shizhe Diao\textsuperscript{2}}, 
 \textbf{Tong Zhang\textsuperscript{1}} 
\\
\small{
    \textsuperscript{1}University of Illinois Urbana-Champaign, 
    \textsuperscript{2}NVIDIA
}
\\
 \small{
    \{ruidaw, ruip4, pw29\}@illinois.edu, shizhe.diao@gmail.com, tozhang@illinois.edu
 }
}

\begin{document}
\maketitle

\maketitle
\begin{abstract}

Enhancing the formal math reasoning capabilities of Large Language Models (LLMs) has become a key focus in both mathematical and computer science communities in recent years. While significant progress has been made in using state-of-the-art Auto-Regressive (AR) LLMs for formal theorem proving, these models suffer from inherent limitations. Their next-token prediction generation methods may yield suboptimal performance due to the challenges of long-range coherence and the compounding of errors over long sequences. Recent advancements in diffusion LLMs (dLLMs), which generate text through iterative denoising of a multi-token block, offer a promising alternative. 
However, the application of dLLMs to formal mathematics, where maintaining long-range coherence is critical, remains largely understudied. 
To address the challenges above, we propose \method, to the best of our knowledge, the first framework to train and apply dLLMs for formal theorem proving. Our frameworks contain training and inference methods for two models. The first one is \textit{dLLM-Prover-7B}, which performs whole-proof writing with long-range coherent tactic usage. 
The second one is \textit{dLLM-Corrector-7B}, which is a novel large block diffusion-based correction model. 
It leverages the in-filling capabilities of dLLMs to perform local proof correction using bi-directional information. 
Extensive experiments demonstrate that \method relatively significantly outperforms the AR LLM baseline trained under the same dataset.
\method achieves an absolute improvement of \textbf{1.61\%} on ProofNet-Test and \textbf{6.14\%} on MiniF2F-Test benchmarks compare to the baseline. 
Notably, \method successfully resolves one IMO problem that more advanced thinking model DeepSeek-Prover-V2-7B~\cite{ren2025deepseek} could not solve, showcasing the unique advantage of dLLMs in formal theorem proving.
\end{abstract}
\section{Introduction}\label{sec:intro}
Building machine-learning systems capable of performing human-level reasoning based on rigorous logical rules has always been considered a fundamental goal for artificial intelligence~\cite{wang2024theoremllama}. This capability is often evaluated through the derivation of complex formal mathematical proofs, where maintaining long-range coherence is essential~\cite{yang2024formal}. However, the inherent ambiguity in natural language makes it challenging to formally verify the intermediate reasoning steps. 

To provide a trustworthy foundation for mathematical reasoning, multiple works have developed verifiable languages based on different theoretical foundations.
Some apply the dependent type language, such as Lean~\cite{de2015lean, moura2021lean} and Coq~\cite{coq1996coq}, which enforce type safety to ensure logical consistency. Others employ higher-order logic, such as Isabelle~\cite{paulson1994isabelle} and HOL~\cite{harrison2009hol}. Proving both types of systems allows explicit verification of every internal step, regardless of its complexity. 
The formal verification minimizes errors and prevents the hallucinations and logical inconsistencies often observed in natural language-based reasoning.

\begin{figure*}[t]
    \vspace{-0.1in}
    \centering
    \includegraphics[width=0.99\linewidth]{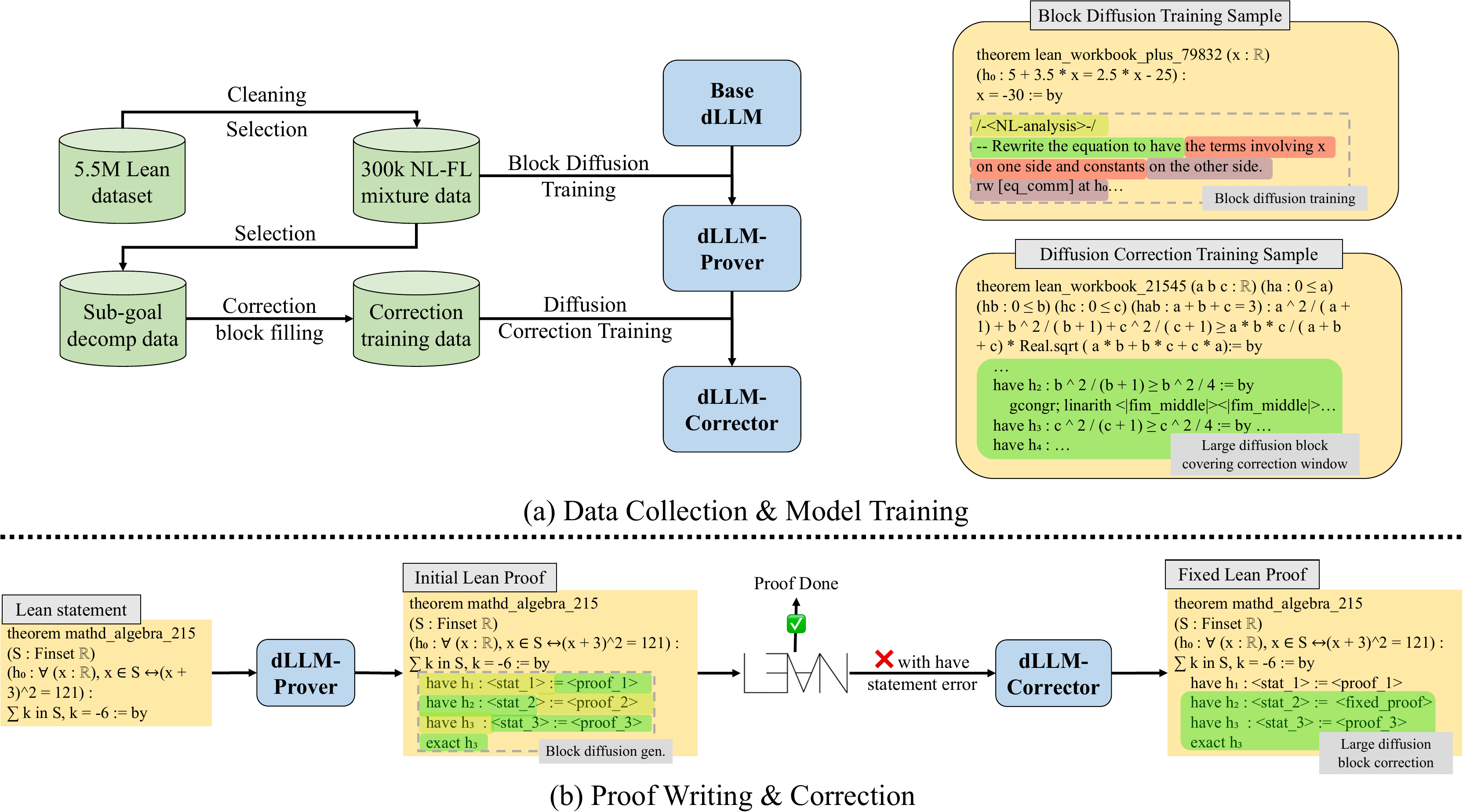}
    \caption{
        \method framework: (a) Data Collection and Model Training: We first collect a 5.5M Lean SFT dataset from previous works, then perform cleaning and selection to obtain 300k natural language (NL)-formal language (FL) mixture data to fine-tune \textit{Fast-dLLM-V2-7B} into \textit{dLLM-Prover-7B}. Subsequently, we select data with subgoal decomposition and perform block filling to formulate data for corrector training. The \textit{dLLM-Corrector-7B} is fine-tuned by the dataset using Large-block correction training. (b) Proof Writing \& Correction: The prover model firstly generates the complete proof for the Lean4 theorems using block-diffusion generation. If the verification fails but the skeleton is correct, the corrector model applies large diffusion block correction to adjust the subgoal proofs based on bi-directional information.
    }
    \label{fig:main}
    \vspace{-0.3in}
\end{figure*}

With the rapid development of Large Language Models (LLMs), their significant potential in formal reasoning using Lean4~\cite{moura2021lean} has become a key area of research. Recent efforts focus on generating large-scale datasets~\cite{ying2024lean, wu2024lean, xin2024deepseek1, wang2024theoremllama} and training models using methods ranging from standard Supervised Fine-Tuning (SFT)~\cite{wang2024theoremllama, lin2025goedel} to advanced Reinforcement Learning (RL)~\cite{ren2025deepseek, wang2025kimina, lin2025goedel2}. 
Latest research proposes multi-agent methods, from recursive decomposition of subgoals~\cite{varambally2025hilbert}, to agentic RL methods~\cite{wang2025gar, chen2025seed} for formal theorem proving. However, most, if not all, previous works on exploring LLMs' usage in formal theorem proving are based on traditional Auto-Regressive (AR) LLMs. The AR models perform next-token prediction in a strict left-to-right manner. It has inherent problems, including potential exponential error accumulation~\cite{dziri2023faith} and a lack of long-range coherence~\cite{ye2025dream}. Furthermore, the AR models are naturally unable to perform in-filling proof correction. These make AR models have natural flaws for formal reasoning. 

On the other hand, diffusion LLMs (dLLMs) present a promising alternative in language generation by reformulating the task as an iterative denoising process of the multi-token block. 
Recent studies~\cite{nie2025large, ye2025dream, wu2025fast, bie2025llada2} have demonstrated their potential in coding tasks, supported by efficient training and inference frameworks ~\cite{wu2025fast2}. While dLLMs naturally benefit from long-range coherence and bi-directional awareness in generation, their application to formal reasoning remains largely understudied.

To address the limitations of existing methods, we propose \method, to the best of our knowledge, the first framework for training and applying dLLMs for formal theorem proving. The overview of our framework can be found in Figure~\ref{fig:main}. 
\method contains the training and inference methods of two 7B-parameter models that work synergically together, namely \textit{dLLM-Prover-7B} and \textit{dLLM-Corrector-7B}. 
In the training stage of \method, we collect 5.5 million records of theorem proving data from previous works, clean and select 300k records of SFT data. It is applied to fine-tune Fast-dLLM-V2-7B into \textit{dLLM-Prover-7B} to enhance its whole-proof writing capability. Subsequently, we introduce a novel Large Block Correction training method, enabling the model to perform local in-filling correction of formal proof with bidirectional information awareness. During inference, the \textit{dLLM-Prover-7B} firstly generates proofs based on the theorem statement. If the initial proof is wrong but the proof skeleton is correct, we apply \textit{dLLM-Corrector-7B} using large diffusion block generation to refine the proof locally.

We summarize our contribution as follows: 
(1) To the best of our knowledge, we present \method, the first framework for fine-tuning and applying the dLLMs for formal theorem proving. 
(2) We introduce the novel large-block correction training and inference method that leverages the bi-directional information to perform in-filling correction. 
(3) Through extensive experiments, we demonstrate that \method relatively significantly outperforms AR LLMs trained on the same dataset, achieving a \textbf{1.61\%} improvement on ProofNet-Test and a \textbf{6.14\%} improvement on MiniF2F-Test under pass@32. Notably, \method successfully solves one IMO problem that the more advanced DeepSeek-Prover-V2-7B~\cite{ren2025deepseek} could not solve.

Our study highlights the unique advantages of dLLMs in theorem proving, and (as far as we know), as the first work in such field, we plan to release all training code, models, and datasets presented in this work.
\section{Methodology}\label{sec:meth}

In this section, we introduce the details of the \method framework. It is designed to leverage the long-range coherence and in-filling correction capabilities of diffusion LLMs (dLLMs) for formal theorem proving. We first present the preliminaries of dLLMs in Section~\ref{meth:pre}. Subsequently, Section~\ref{meth:data} details the data preparation and Section~\ref{meth:train} presents model training. Finally, we present the proof writing and correction method that applies the trained model in Section~\ref{meth:write}.

\subsection{Preliminaries}\label{meth:pre}

Consider a token sequence $\bm{x} = \{x_1, x_2, \cdots, x_L\}$ with length $L$, traditional Auto-regressive (AR) LLMs generate text sequentially following a stright left-to-right manner by modeling the conditional distribution $\mathbb{P}_\theta\left(x_i | \bm{x}[0:i - 1]\right)$, where $\theta$ is the AR model's parameters. These models are trained to minimize cross-entropy loss of predicted tokens based only on previous tokens.

Diffusion LLMs, on the other hand, approach text generation differently by introducing a time step $t \in [0, 1]$ and adding noise to create $\bm{x}^{(t)}$, where tokens are masked independently with probability $t$. The model $\mathbb{P}_\omega (\bm{x}^{(0)} | \bm{x}^{(t)})$ iteratively predicts the original tokens from the noisy input. The training objective for dLLMs is defined as: $\mathcal{L}(\theta) = - \mathbb{E}_{t, \bm{x}^{(0)}, \bm{x}^{(t)}} [\sum_{i = 1}^L \mathbb{I}[x_i^{(t)} = \text{[MASK]} ] \cdot \log \mathbb{P}_\omega (x_i^{(0)} | \bm{x}^{(t)})$

In the \method, we apply the block dLLM framework as the base model. Unlike the traditional dLLMs, the sequence is divided into blocks of size $B$, with bi-directional attention restricted to individual blocks while preserving causal relationships between blocks. During inference, the block dLLMs iteratively decode one block at a time until a stopping condition is reached. This process balances the diffusion generation and the flexible nature of AR models.

\subsection{Data Collection \& Pre-process}\label{meth:data}

\begin{figure*}
\centering
\begin{adjustbox}{max width=0.99\linewidth}
\begin{tcolorbox}
\begin{lstlisting}
theorem lean_workbook_21545_1 (a b c: ℝ) (ha: 0 ≤ a) (hb: 0 ≤ b) (hc: 0 ≤ c) (hab: a + b + c = 3) :  a^2 / (a + 1) + b^2 / (b + 1) + c^2/(c + 1) ≥ a*b*c / (a + b + c) * Real.sqrt (a*b + b*c + c*a):= by
  rw [hab]
  have : 0 ≤ a*b*c := by positivity
  have h₁: a^2 / (a + 1) ≥ a^2 / 4 := by gcongr; linarith
  have h₂: b^2 / (b + 1) ≥ b^2 / 4 := by gcongr; linarith
  have h₃: c^2 / (c + 1) ≥ c^2 / 4 := by gcongr; linarith
  have h₄: a^2 / 4 + b^2 / 4 + c^2 / 4 ≥ a*b*c / 3 * Real.sqrt (a*b + b*c + c*a) := by
    have : a*b*c ≤ 1 := by
      nlinarith [sq_nonneg (a - b), sq_nonneg (b - c), sq_nonneg (c - a)]
    have : Real.sqrt (a*b + b*c + c*a) ≤ 2 := by
      apply Real.sqrt_le_iff.mpr
      constructor
      · positivity
      · nlinarith [sq_nonneg (a - b), sq_nonneg (b - c), sq_nonneg (c - a)]
    nlinarith [sq_nonneg (a - b), sq_nonneg (b - c), sq_nonneg (c - a)]
  nlinarith
\end{lstlisting}
\end{tcolorbox}
\end{adjustbox}
\caption{Example of data with subgoal-decomposition, the \texttt{have h$_i$} at top-level is the decomposed subgoal}\label{fig:subgoal_decomp}
\vspace{-0.15in}
\end{figure*}

This section outlines the data collection and preprocessing steps for fine-tuning \textit{dLLM-Prover-7B} and \textit{dLLM-Corrector-7B} models in \method. While previous works~\cite{wu2024lean, lin2025goedel, dong2025stp, wang2025gar, wang2025kimina} have contributed millions of theorem proof records. The data lacks proper structuring. Specifically, these datasets are often in their raw input-output form and lack a clear separation between FL statements, NL statements, and FL proofs. 

Thus, to uniform the dataset, we extract and standardize the data using rule-based methods to separate NL statements, FL statements, and FL proofs. Additionally, proof data from STP~\cite{dong2025stp} lacks NL components and is used as syntax understanding data for Lean. In total, we obtained 5,595,798 records of code-completion proofs and sampled 300k records to create the SFT dataset for prover. We maintain an approximate ratio of 1:2 between pure Lean proofs and those with NL annotations and comments. Following the DeepSeek-Prover~\cite{xin2024deepseek1} standard, we format the data in a code-completion style without adopting a chat template. Examples of the training data are provided in Section~\ref{data:sft}.

We then prepare the data for the training corrector by sampling from the 300k data that has subgoal decomposition in its proof.
For a Lean proof, subgoals are defined as internal statements and their proofs in the theorem proof are marked by the \texttt{have} keyword, as illustrated in Figure~\ref{fig:subgoal_decomp}. 
We organize the top-level subgoal proof into token blocks of size 256 by filling the proof with a placeholder \texttt{<|fim\_middle|>} token. The data trains the model to fill placeholder token when a block of proof is finished. Additionally, we only consider one to fill in the block at a time for a proof containing multiple subgoals. Following the above preprocessing, we construct a dataset containing 300k records of SFT data for prover training and 128k records for corrector training.

\subsection{Prover \& Corrector Training}\label{meth:train}

After obtaining the dataset, we fine-tune the \textit{dLLM-Prover-7B} and \textit{dLLM-Corrector-7B} models. For the prover model, we start with the base model Fast-dLLM-V2-7B~\cite{wu2025fast2} and employ a standard text-to-text training framework under the Lean4 code-completion task. To ensure the stability of training, we maintain the diffusion block size to be 32, which is consistent with the original model. Additionally, we implement curriculum data sorting~\cite{wang2024theoremllama}, which organizes the training by increasing order of proof complexity based on proof length. This approach enables the model to first learn simpler proofs before progressing to more complex ones, thereby stabilizing the loss curve. Following the above training process on the 300k SFT dataset, the \textit{dLLM-Prover-7B} becomes proficient in formal theorem proving. Such training enables the model to write formal proofs with better long-range coherence in the sense of planning and tactic usage through block diffusion generation.

Subsequently, we further leverage the bi-directional information understanding of dLLMs by proposing a large-block training method for the corrector.
Under such a training method, we extend the diffusion block size from 32 to 512 and apply loss to both the subgoal proof block and the placeholder token. It aims to train the model to write local corrections and fill the rest of the block with placeholder tokens. We apply such a training method to \textit{dLLM-Prover-7B} and obtain the \textit{dLLM-Corrector-7B}. 
This enables the corrector to utilize information from the extended diffusion block to perform in-filling corrections on subgoals by leveraging bi-directional information.

\subsection{Proof Writing \& Correction}\label{meth:write}

Using the \textit{dLLM-Prover-7B} and \textit{dLLM-Corrector-7B} trained from above steps, \method is able to perform both whole-proof generation and error correction with long-range coherence. During the whole-proof writing stage, the \textit{dLLM-Prover-7B} generates proofs based on the NL and Lean4 statement of the theorem. This process is based on the proof generation method from DeepSeek-Prover-V1~\cite{xin2024deepseek1}.

If the initial proof by the prover fails the Lean verification but all the top-level subgoal statements and their usages are correct, we perform a large-block correction using the corrector.
The correction process begins with deleting the proofs in the subgoal that causes the error and replacing them with a 256-token sequence of diffusion generation mask \texttt{<|MASK|>}.
The \textit{dLLM-Corrector-7B} then refines these masked sections using 512-length diffusion blocks, such that a large block can leverage both prefix and suffix context to perform in-filling corrections. To encourage more creative and accurate proof writing, we set the generation temperature to 1.2. Additionally, the confidence of one token to be denoised is increased to 0.95 to ensure the robustness of the generation. The input-output example for \textit{dLLM-Corrector} can be found in Appendix~\ref{appendix:input}.

Through the above inference process, we are able to perform long-range coherent proof generation that surpasses the limitations of AR generation and effectively fixes local errors through in-filling corrections.
\section{Experiments}\label{sec:exp}

We evaluate \method on Lean4 theorem proving through extensive experiments on MiniF2F-Test~\cite{zheng2021minif2f} and ProofNet-Test~\cite{azerbayev2023proofnet} benchmarks. Section~\ref{exp:setup} and~\ref{exp:implementation} describe the experiment setup and implementation details. Subsequently, Section~\ref{exp:results} reports the main results of our experiment. Furthermore, Section~\ref{exp:loss} provides further analyses in the training-loss study; Section~\ref{exp:abl} analyzes the performance after dropping out the key component, and Section~\ref{exp:case} offers detailed case studies.

\subsection{Experiment Setup}\label{exp:setup}

\subsubsection{Dataset and Task}\label{setup:data}

\begin{table*}
    \centering
    \resizebox{\textwidth}{!}{
    \tiny
    \begin{tabular}{ccccc}
        \toprule
        \textbf{Dataset}    & \textbf{Data Number}  & \textbf{Qwen-2.5-Lean-SFT-7B}    & \textbf{Diffusion-Proof}    & \textbf{Improvement} \\
        \midrule
        \textbf{ProofNet-Test}          & 186   & 5.91\%        & \textbf{7.53\%}       & 1.61\% \\
        \midrule
        \textbf{MiniF2F-Test}           & 244   & 43.85\%       & \textbf{50.00\%}      & 6.14\% \\
        \midrule
        \textit{MiniF2F-Test by problem type} \\
        \textbf{IMO}                    & 20    & 5.00\%        & \textbf{15.00\%}      & 10.00\% \\
        \textbf{AIME}                   & 15    & 33.33\%       & 33.33\%               & 0.00\% \\
        \textbf{AMC}                    & 45    & 22.22\%       & \textbf{26.67\%}      & 4.44\% \\
        \textbf{Algebra}                & 88    & 55.68\%       & \textbf{67.05\%}      & 11.36\% \\
        \textbf{Number Theory}          & 68    & 58.82\%       & \textbf{63.24\%}      & 4.41\% \\
        \textbf{Induction}              & 8     & 25.00\%       & 25.00\%               & 0.00\% \\
        \bottomrule
    \end{tabular}
    }
    \vspace{-0.05in}
    \caption{Main experiment result of \method under MiniF2F-Test~\cite{zheng2021minif2f} and ProofNet-Test~\cite{azerbayev2023proofnet} with pass@32 metric and 32 corrections for each valid theorem}
    \label{tab:main}
    \vspace{-0.2in}
\end{table*}

We evaluate the formal reasoning capabilities of the \method framework by MiniF2F-Test~\cite{zheng2021minif2f} and ProofNet-Test~\cite{azerbayev2023proofnet} benchmarks. These datasets are frequently employed in major studies on the LLM formal reasoning field~\cite{xin2024deepseek1, ren2025deepseek, wang2024theoremllama, dong2025stp, lin2025goedel2, wang2025gar}. The MiniF2F-Test datasets comprise 244 math problems formalized in Lean4. Its difficulty ranges from high-school competition problems to elementary undergraduate-level problems. The problems include formalized competition problems from IMO, AIME, and AMC, a subset of the Math-500 benchmark~\cite{lightman2023let}, and hand-crafted problems with a similar level of difficulty. 
On the other hand, the ProofNet-test contains 186 theorems derived from college textbooks, covering advanced topics such as real and complex analysis, linear algebra and topology.

The task for the LLMs is to generate formal proofs for theorems based on their NL and Lean4 statements using direct code-completion. We do not allow Long CoT thinking process because both \method and baselines does not support such long context.

\subsubsection{Baselines \& Evaluation Metric}\label{setup:baseline}

The primary goal of this study is to evaluate the potential of dLLMs in formal theorem proving, particularly their capability to achieve long-range coherent tactics usage. 
Thus, for a fair comparison, we use Qwen-2.5-Instruct-7B~\cite{qwen2025qwen25}, which is fine-tuned on the same training dataset as \textit{dLLM-Prover-7B} and employs the same text-to-text training method. 
The SFT version is named \textit{Qwen-2.5-Lean-SFT-7B} in our experiment. We select this baseline because Qwen-2.5-Instruct-7B is the base model for Fast-dLLM-V2-7B~\cite{wu2025fast2}, and neither model undergoes additional post-training focused on mathematics or Lean4 coding. Therefore, both models share a comparable foundation in formal reasoning. It makes the performance improvement of \method a strong indicator of the effectiveness of the diffusion-based framework. 

We emphasize that the primary goal of this study is to investigate whether dLLMs offer architectural advantages over AR models for formal reasoning, rather than achieving state-of-the-art performance. Therefore, we deliberately control for confounding factors and compare only the generation paradigm. For reference, contemporary SFT-only provers like DeepSeek-Prover-V1.5 achieve approximately 48.2\% on MiniF2F with larger-scale training, suggesting the meaningfulness of our framework.

To measure the performance, we adopt the standard pass@32 metric applied in almost all previous studies. The details of the evaluation metric can be found in Appendix~\ref{appendix:eval}.

\subsection{Implementation Details}\label{exp:implementation}

This section outlines the implementation details for the prover, corrector, and baseline models. The prover and baseline models are fine-tuned on the SFT dataset described in Section~\ref{meth:data} using a learning rate of 1E-5 and a context length of 2,048 tokens. On the other hand, the corrector is trained with a reduced learning rate of 5E-6. All models share the same other training configuration, employing a learning rate scheduler with 3\% warm-up steps. Training is conducted on 4-card H100 GPUs with a global batch size of 64. Each model requires approximately one day of training. For detailed training and inference setup, please refer Appendix~\ref{appendix:imp}.

\subsection{Results}\label{exp:results}

The main evaluation results is presented in Table~\ref{tab:main}. It shows that \method achieves an accuracy rate of 50.00\% on MiniF2F and 7.53\% on ProofNet. The result surpasses the \textit{Qwen-2.5-Lean-SFT-7B} baseline by 6.14\% and 1.61\%, respectively. The baseline records 43.85\% on MiniF2F and 5.91\% on ProofNet.

A detailed analysis of MiniF2F results indicates that \method outperforms the baseline across most problem types. Notably, it demonstrates relatively significant gains in challenging IMO and AMC problems, solving two additional problems in each category. Furthermore, comparably significant improvements are also observed in Algebra and Number Theory. It highlights the framework's enhanced formal reasoning capabilities in both fields. On the more challenging ProofNet benchmark, despite the lower absolute improvements, \method also shows clear progress over the baseline. We also compare \method with broader baselines and the performance of \method on harder problems that require longer proofs. Detailed results are provided in Appendix~\ref{add_exp:baseline} and ~\ref{add_exp:long_proof}.

\subsection{Validation Loss Study}\label{exp:loss}

In this section, we investigate why dLLMs outperform AR models in formal proof writing based on the validation loss. Using the 190 correct Lean theorem proofs from MiniF2F-Test completed by DeepSeek-Prover-V2~\cite{ren2025deepseek}. We compare the cross-entropy loss of different models under the pure causal mask. For dLLMs, this comparison is achieved by setting the diffusion block size to 1. We evaluate four models: the fine-tuned dLLM (\textit{dLLM-Prover-7B}), the fine-tuned AR model (\textit{Qwen-2.5-Lean-SFT-7B}), the base dLLM (Fast-dLLM-V2-7B), and the base AR model (Qwen-2.5-Instruct-7B). The validation loss distribution is presented in Figure~\ref{fig:valid_loss}, whith detailed plots available in Appendix~\ref{appendix:loss}.

\begin{figure}[h]
    \centering
    \includegraphics[width=0.95\linewidth]{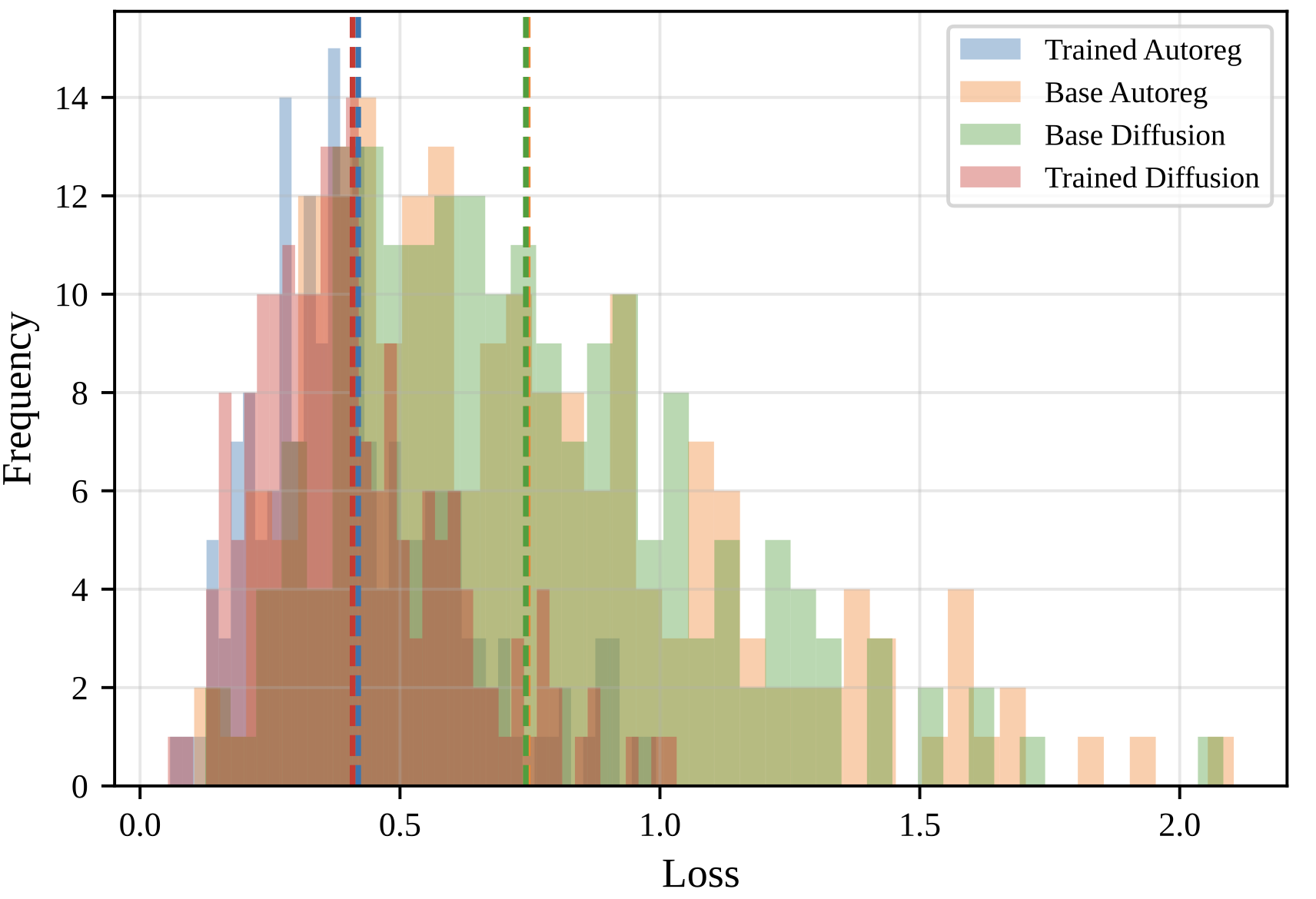}
    \caption{Validation loss for trained dLLM, trained AR LLM, base dLLM, and base AR LLM}
    \label{fig:valid_loss}
    \vspace{-0.3in}
\end{figure}

The results reveal that the two base models exhibit nearly identical loss distributions, as well as two fine-tuned models. The statistical analysis shows that the Pearson correlation for SFT models is 0.9846 and for base models, it is 0.9838. The p-value for both correlations is under $10^{-100}$. This demonstrates that, in terms of causal attention loss, the models within each category are almost equivalent. However, the fine-tuned dLLM relatively significantly outperforms the SFT AR model across all benchmarks. We attribute this advantage to the natural diffusion generation. It features iterative refinement in the diffusion block, enhanced long-range coherence, and bi-directional information awareness. The loss comparison highlights the superior capability of dLLMs in formal reasoning.

\subsection{Ablation Study in Correction Process}\label{exp:abl}

This section presents an ablation study to evaluate the impact of the correction process in the \method framework. The corrector contributes to solving 4 out of 122 problems in the MiniF2F-Test, namely \texttt{imo\_1962\_p2}, \texttt{amc12b\_2002\_p7}, \texttt{mathd\_algebra\_215}, and \texttt{mathd\_numbertheory\_521}. It results in a 1.64\% improvement in the benchmark. These problems typically involve complex subgoal decomposition and require the corrector to further split objectives into multiple cases to complete the proof. The detailed case study is presented in Section~\ref{case:corrector}. On the other side, even without the corrector, the \textit{dLLM-Prover-7B} still outperforms the baseline by 4.51\% on MiniF2F, validating the effectiveness of dLLMs for formal reasoning. 

However, on the ProofNet-test dataset, the corrector's local proof fix proves no additional theorems. This is because such a dataset focuses more on advanced knowledge over complex subgoal division or long-range reasoning. With its inherent difficulty, such a dataset is naturally not suitable for a corrector's local proof fix. We further perform the analysis using the AR model and \textit{dLLM-Prover-7B} to replace the corrector and to compare in-place correction with top-level correction; the results are shown in Appendix~\ref{add_exp:dual}.

\subsection{Case Study}\label{exp:case}

This section presents a detailed case study to evaluate the qualitative performance of \method. We only present analysis here with additional examples provided in Appendix~\ref{appendix:example}.

\subsubsection{Theorems proved by corrector}\label{case:corrector}

This section analyzes the performance of \textit{dLLM-Corrector-7B} in successfully performing local theorem proofs based on a skeleton generated by \textit{dLLM-Prover-7B}. We select two representative cases, namely \texttt{mathd\_algebra\_215} and \texttt{mathd\_numbertheory\_521}, for the case study. 

In the case of \texttt{mathd\_algebra\_215}, both the initial failed proof (Figure~\ref{fig:fail_alg_215}) and the corrected proof (Figure~\ref{fig:succ_alg_215}) share the same global structure. The proof begins by establishing the elements of the set $S$ and then demonstrating that these elements satisfy the given conditions. However, the original proof struggles to apply tactics for handling complex algebraic computations. On the contrary, the corrector generates the entire proof in a single diffusion block, enabling a more strategic planning. The process begins with generating the internal \texttt{have} decomposition, followed by proving each subgoal. It demonstrates the corrector's capability to perform long-range coherent reasoning and tactic usage. Additionally, the final \texttt{intro h} steps mirror the type refinement seen in the latter parts of the proof. It further highlights the model's bi-directional context awareness, which enables it to complete the proof based on suffix information.

In the case of \texttt{mathd\_numbertheory\_521}, the original proof fails due to incorrect usage of \texttt{linarith} and \texttt{rw} tactics. These tactics are less effective in the domain of natural numbers compared to \texttt{omega}, which is used in the corrected proof. Leveraging the corrector's long-range generation capability and large attention block, it identifies the appropriate tactic and applies it correctly. Notably, the proof of \texttt{h$_6$} is generated before \texttt{h$_4$}, and the bi-directional awareness of the diffusion corrector allows the model to learn from the suffix generation of the same block directly. This further proves the advantages of the corrector's large diffusion block in capturing and utilizing contextual information effectively.

\subsubsection{Compare to AR baseline model}\label{case:AR}

We analyze two representative cases where \textit{dLLM-Prover-7B} succeeds while the AR baseline fails, namely \texttt{imo\_1983\_p6} and \texttt{mathd\_algebra\_188}. The complete examples are provided in Appendix~\ref{example:ar}.
In the case of \texttt{imo\_1983\_p6}, both models attempt to prove the inequality by supplying complex conditions to the tactic \texttt{linearity}. The dLLM's block diffusion generation enables it to take a more comprehensive strategy, providing suitable input for the theorem \texttt{mul\_nonneg}. In contrast, the AR model fails to provide adequate inputs, rendering the wrong theorem proof. In the \texttt{mathd\_algebra\_188}, the AR model applies incorrect tactics in both subgoal proofs, leaving the theorem incomplete. On the other hand, the dLLM employs a more effective approach. It successfully solves the subgoals. This comparison highlights the dLLM's ability to perform coherent planning, a capability that the AR models lack.

\subsubsection{Compare to DeepSeek-Prover-V2}\label{case:DS}

One surprising result from our experiments is that \method successfully completes the proof for \texttt{imo\_1962\_p2}. The problem fails the DeepSeek-Prover-V2-7B~\cite{ren2025deepseek} under pass@32 with Long CoT reasoning. The detail of this example is presented in Appendix~\ref{example:ds}. As shown in Figure~\ref{fig:ds_anal_imo_1962}, despite DS-Prover providing sufficient natural language analysis during the Long CoT process, it lacks a detailed proof plan and coherent tactic usage. This deficiency leads to repeated self-corrections in the natural language plan without providing additional insight into Lean beyond direct code writing. Although the model easily proves the condition $-1 \leq x$. It fails on the second case, by providing the wrong direction of inequality in \texttt{h$_{19}$}, which collapses the entire proof. In contrast, the corrector in \method demonstrates a more robust and coherent tactic usage. It successfully addresses the long-range dependencies required for this proof. By leveraging its large diffusion block, the corrector can have better planning and execution of complex reasoning steps. This highlights the potential of dLLMs in formal theorem proving, particularly for tasks requiring intricate tactic planning.
\section{Related Work}\label{sec:relat}

\subsection{LLMs For Formal Theorem Proving}\label{relat:fl}

In recent years, LLMs have gained prominence in formal theorem proving with Lean4~\cite{moura2021lean} emerging as a widely used environment for formal verification tasks. Multiple works have contributed millions of records of theorem statements and proofs~\cite{wang2024theoremllama, wu2024lean, lin2025goedel, dong2025stp}, establishing a robust foundation for the domain. 

In the sense of model training, early advancements in training LLMs for theorem proving introduced SFT frameworks, including Expert Iteration~\cite{polu2022formal}, Re-Prover~\cite{yang2024leandojo}, TheoremLlama~\cite{wang2024theoremllama}, DeepSeek-Prover-V1~\cite{xin2024deepseek1}, BFS-Prover~\cite{xin2025bfs}, and Goedel-Prover-V1~\cite{lin2025goedel}. Subsequent developments leveraged RL with verifier rewards, leading to the reasoning systems such as MA-LoT~\cite{wang2025ma}, Kimina-Prover~\cite{wang2025kimina}, Goedel-Prover-V2~\cite{lin2025goedel2}, and DeepSeek-Prover-V2~\cite{ren2025deepseek}. More recently, agentic-RL has been actively explored for multi-agent proving systems, leading to frameworks like GAR~\cite{wang2025gar} and Seed-Prover-V1.5~\cite{chen2025seed}, which further improved the performance in theorem proving. Despite these significant strides, most, if not all, existing approaches rely on AR LLMs, leaving the dLLM largely under-explored in the context of formal reasoning. 

\subsection{Diffusion LLMs}\label{relat:dllm}

Diffusion Large Language Models (dLLMs) have emerged as a promising alternative to Auto-Regressive (AR) models for text generation. Unlike AR models, which generate tokens sequentially from left to right, dLLMs generate text using a denoising process. It progressively refines the masked tokens through multiple diffusion steps~\cite{nie2025large, ye2025dream}. This iterative approach inherently supports long-range coherence, a feature often lacking in AR architectures.

Early foundational efforts, such as LLaDA~\cite{nie2025large} and Dream~\cite{ye2025dream}, demonstrated the potential of this paradigm. Subsequent advancements, including efficient inference frameworks like dInfer~\cite{ma2025dinfer} and Fast-dLLM family~\cite{wu2025fast} have substantially improved the practical deployment of dLLMs through innovations in KV-cache management and system-level optimizations. Additionally, block diffusion techniques~\cite{wu2025fast2} extend diffusion models beyond fixed-length generation, enabling greater flexibility. LLaDA 2.0~\cite{bie2025llada2} scales dLLMs to 100B parameters and has showcased the viability of these architectures at the frontier of language modeling. Despite their growing application in downstream tasks~\cite{you2025llada, dong2025llada, shi2025llada}, applying dLLMs for formal reasoning is still an under-explored field. This gap highlights an opportunity for further research into applying dLLMs to tasks requiring rigorous logical coherence.
\section{Conclusion}\label{sec:conc}

This paper presents \method, a comprehensive training and inference framework for applying diffusion LLM (dLLM) for formal theorem proving. \method aims to address limitations of Auto-Regressive (AR) models, such as their inherent lack of long-range coherence in proof generation and inability to perform bi-directional aware, in-filling proof corrections. \method tries to overcome these challenges by integrating two key components: \textit{dLLM-Prover} and \textit{dLLM-Corrector}. The prover employs block diffusion training and generation to enable long-range coherent tactic usage and improve long-range proof planning. The corrector, a large-block diffusion model, performs in-filling proof corrections using bi-directional context, allowing the system to tackle more complex problems. Extensive experiments demonstrate that \method framework outperforms the AR baseline trained under the same dataset by \textbf{6.14\%} in MiniF2F-Test and \textbf{1.61\%} in ProofNet-Test under pass@32. Notably, with the integration of the corrector, \method successfully solves an IMO problem that much stronger DeepSeek-Prover-V2-7B cannot handle. Beyond theorem proving, \method provides a promising approach by highlighting the unique advantages of dLLMs in tasks that require long-range coherence and precise reasoning.

\section*{Limitations}

Despite \method presenting promising results, there are areas that warrant further exploration as the application of dLLMs in formal reasoning and other downstream tasks is still in its nascent stages. First, this work is conducted with limited computational resources, which restricts the training scale below the state-of-the-art open-source AR provers such as Goedel-Prover-V2~\cite{lin2025goedel2} and DeepSeek-Prover-V2~\cite{ren2025deepseek}. Second, the current base model's limited Long CoT capability prevents a comprehensive exploration of dLLM's potential for long thinking in theorem proving. Third, this work focuses only on Lean4. Although benchmarks such as MiniF2F include partial formalizations in other theorem provers, training a dLLM prover requires large-scale proof corpora, verification infrastructure, and model-specific preprocessing. Comparable resources are currently much stronger for Lean4 than for theorem provers such as Isabelle in our setting. Finally, this study focuses primarily on the empirical improvements and does not delve into the theoretical foundations of dLLMs. These aspects provide valuable opportunities for future research to scale training further, develop models with enhanced reasoning capabilities, and establish a theoretical framework to further understand and advance the usage of dLLMs in formal reasoning tasks.

\bibliography{custom}

\newpage
\appendix\label{sec:appendix}

\section{Discussion}

\subsection{AI usage}
This work utilized Copilot to assist with code writing and OpenAI's models to correct grammatical issues in the paper. All the ideas of the paper are original.

\subsection{Potential Risk}
This work focuses on applying a new form of model beyond AR LLMs to physics theorem proving tasks; there is no foreseeable potential risk for this paper.

\subsection{Discussion on Data}
The data used in this paper are based on open-source data from previous peer-reviewed works, which adhere to the ethics standards of major conferences, introducing no foreseeable risk of leaking personally identifiable information or offensive content.

\subsection{Discussion on License and Scientific Artifact Use}
The datasets we use are published under the MIT License, and the base model Qwen-2.5 is published under the Apache 2.0 License. When we open-source the model, it follows the Apache 2.0 License, and the dataset follows the MIT License. Additionally, with internal checks, authors agree that the scientific artifact follows the original intention.

\section{Additional Implementation Detail}\label{appendix:imp}

The global batch size is implemented with per-GPU batch size 2 and gradient accumulation 8. Training \textit{dLLM-Prover-7B} and \textit{Qwen-2.5-Lean-SFT-7B} each takes approximately 96 H100 GPU-hours, while training \textit{dLLM-Corrector-7B} takes an additional 48 H100 GPU-hours. During evaluation, the complete \method pipeline, including whole-proof generation, Lean verification, and correction, takes about 16 hours on our machine. The AR baseline takes about 24 hours without vLLM optimization.

\section{Additional Experiment Results}\label{appendix:add_exp}

\subsection{Compare with Broader Baselines}\label{add_exp:baseline}

\begin{table*}[t]
    \centering
    \resizebox{0.95\textwidth}{!}{
    \small
    \begin{tabular}{lccc}
        \toprule
        \textbf{Method} & \textbf{Size} & \textbf{Training / Inference Setting} & \textbf{MiniF2F-Test} \\
        \midrule
        TheoremLlama~\cite{wang2024theoremllama} & 8B & AR SFT & 35.7\% \\
        Lean-STaR & 7B & AR SFT / expert iteration & 46.3\% (pass@64) \\
        DeepSeek-Prover-V1~\cite{xin2024deepseek1} & 7B & AR SFT & 46.3\% \\
        DeepSeek-Prover-V1.5~\cite{xin2024deepseek} & 7B & AR SFT + RL & 50.0\% \\
        Kimina-Prover-Preview~\cite{wang2025kimina} & 1.5B & AR SFT + RLVR + Long CoT & 56.2\% \\
        DeepSeek-Prover-V2~\cite{ren2025deepseek} & 7B & AR SFT + RLVR + Long CoT & 70.49\% \\
        \method & 7B & dLLM SFT & 50.0\% \\
        \bottomrule
    \end{tabular}
    }
    \vspace{-0.02in}
    \caption{Contextual comparison with specialized theorem provers. This table is not a controlled comparison because the systems differ in data scale, RL usage, search, and Long CoT reasoning.}
    \label{tab:broader_baselines}
    \vspace{-0.12in}
\end{table*}

In this section, we provide a detailed comparison of provers with different sizes and training methods to better demonstrate the position of \method in the entire prover ecosystem. The results are demonstrated in Table~\ref{tab:broader_baselines}.

From these results, we observe that \method, trained only on 300k data with standard SFT, outperforms AR-based SFT models on larger datasets and achieves results comparable to early-stage RL methods. This demonstrates the potential of diffusion models for formal theorem proving. However, the lack of the RLVR (Reinforcement Learning with Verifiable Feedback) technique and Long CoT reasoning currently limits our model's ability to surpass SOTA AR provers. This result demonstrates that diffusion provers provide a stronger foundation for models and point the future direction of developing better RL methods for the field.

\subsection{Long Proof Subset Comparison}\label{add_exp:long_proof}

\begin{table}[t]
    \centering
    \small
    \resizebox{\linewidth}{!}{
    \begin{tabular}{lcc}
        \toprule
        \textbf{MiniF2F Subset} & \textbf{\method} & \textbf{Qwen-Lean-SFT-7B} \\
        \midrule
        Longest 25\% & 5/54 (9.26\%) & 4/54 (7.41\%) \\
        Longest 50\% & 26/108 (24.07\%) & 20/108 (18.52\%) \\
        \bottomrule
    \end{tabular}
    }
    \vspace{-0.02in}
    \caption{Performance on MiniF2F-Test subsets with the longest released DeepSeek-Prover-V2 proof lengths.}
    \label{tab:long_proof}
    \vspace{-0.12in}
\end{table}

To evaluate whether the gain is concentrated in problems requiring longer formal derivations, we sort MiniF2F-Test problems by the length of released DeepSeek-Prover-V2 proofs and evaluate performance on the longest subsets. As shown in Table~\ref{tab:long_proof}, \method solves 5/54 problems on the longest 25\% subset and 26/108 problems on the longest 50\% subset, compared with 4/54 and 20/108 for the AR baseline. The larger gap on the longest half of the benchmark supports our hypothesis that block diffusion generation is particularly useful when proof construction requires longer-range tactic consistency.

More surprisingly, \method successfully solves one problem (\texttt{imo\_1962\_p2}) in MiniF2F-Test that the more advanced DeepSeek-Prover-V2-7B~\cite{ren2025deepseek}, equipped with Long CoT, fails to prove under pass@32. This further highlights the unique advantage of dLLMs in generating long-range coherent proofs that even more sophisticated reasoning models struggle to complete.

\subsection{Additional Ablation Study on Correction Process}\label{add_exp:dual}

\begin{table}[t]
    \centering
    \small
    \resizebox{\linewidth}{!}{
    \begin{tabular}{lc}
        \toprule
        \textbf{Model / Correction Setting} & \textbf{MiniF2F-Test} \\
        \midrule
        \textit{dLLM-Corrector-7B} as unified model & 40.98\% \\
        \textit{dLLM-Prover-7B} only & 48.36\% \\
        \textit{dLLM-Prover-7B} + \textit{dLLM-Corrector-7B} & 50.00\% \\
        \midrule
        AR line-level rewrite correction & +1 problem \\
        dLLM in-place correction & +2 problems \\
        dLLM top-level correction & +4 problems \\
        \bottomrule
    \end{tabular}
    }
    \vspace{-0.02in}
    \caption{Ablations for dual-model design and correction strategy on MiniF2F-Test. The lower block reports additional problems solved by each correction method.}
    \label{tab:corrector_abl}
    \vspace{-0.12in}
\end{table}

We further test whether a unified model can replace the proposed dual-model design by using \textit{dLLM-Corrector-7B} for both whole-proof writing and correction. This pure-corrector setup reaches only 40.98\% on MiniF2F-Test, compared with 50.00\% for the proposed pipeline. The result indicates that large-block correction training improves local infilling behavior but harms first-attempt whole-proof generation, supporting the separation between global proof writing and local correction.

We also compare the top-level correction design with an in-place correction baseline that removes the erroneous line and regenerates the remaining proof inside the same subgoal. In-place correction solves 2 additional MiniF2F-Test problems, while top-level correction solves 4. This suggests that regenerating the entire top-level subgoal proof gives the corrector more freedom to revise the local proof trajectory, whereas in-place correction remains constrained by earlier potentially misleading tactic choices. Finally, we equip the AR baseline with a correction mechanism by asking \textit{Qwen-2.5-Lean-SFT-7B} to rewrite from the line before the first error. This corrects only one problem, \texttt{mathd\_algebra\_275}, which is already solved by \textit{dLLM-Prover-7B} in whole-proof generation.

\section{Error Analysis}\label{appendix:error}

We analyze seven representative failed proofs to identify common limitations of \method. Four examples are caused by overconfident use of arithmetic tactics such as \texttt{linarith} or \texttt{nlinarith}: \path|induction_pord1p1on2powklt5on2|, \path|amc12_2000_p20|, \path|amc12a_2020_p15|, and \path|algebra_2varlineareq_fp3zeq11_3tfm1m5zeqn68_feqn10_zeq7|. In these cases, the generated proof often follows a plausible high-level algebraic plan, but then asks Lean arithmetic tactics to discharge obligations that require missing nonlinear, complex-number, or domain-specific transformations. For example, in \path|amc12a_2020_p15|, the model attempts to derive complex-valued polynomial consequences using arithmetic tactics that are not sufficient for the goal.

One example, \path|amc12a_2021_p19|, is caused by invalid tactic usage. The model invokes simplification tactics with theorem combinations that do not make progress in Lean, indicating that the tactic pattern is locally plausible but not applicable to the current goal state.

The remaining two examples, \path|imo_1981_p6| and \path|amc12b_2020_p13|, exceed the current model capacity and show repetitive or incomplete proof endings. These failures occur on problems requiring substantially longer decomposition or specialized identities, suggesting that stronger Long CoT reasoning, verifier-guided self-reflection, or additional difficult-proof data may be needed. 

\section{Training data examples}\label{appendix:data}

\subsection{dLLM-Prover SFT Data Example}\label{data:sft}

The training input-output example for \textit{dLLM-Prover} is presented in Figure~\ref{fig:inout_sft}.

\subsection{dLLM-Corrector Data Example}\label{data:corrector}

The input and output example for \textit{dLLM-Corrector} is presented in Figure~\ref{fig:inout_corrector}. Additionally, we demonstrate the extent to which a single 512-token diffusion block can cover in the example. From the data example, we can observe that the target block for correction is \texttt{h$_3$}, the training diffusion block covers both prefix and suffix. Through training on the presented data, the corrector model learns to write proofs for subgoals with awareness of bi-directional information.

\section{Input-output Examples for corrector}\label{appendix:input}

The input example for the corrector is presented in Figure~\ref{fig:in_gen_corrector}, and the output example is in Figure~\ref{fig:out_gen_corrector}. From the example, we can see that the input sequence replaces the original wrong proof in the \texttt{h$_1$} with a 256 mask token and leaves the corrector to perform in-filling correction. For the output example, we can see the model successfully writes the proof for the decomposed subgoal and learns to fill the extra space for the diffusion block with placeholder tokens. This indicates the success of our training in the corrector, enabling it to perform in-filling correction and complete the proof with placeholder tokens.

\section{Evaluation Metrics}\label{appendix:eval}

The pass@32 evaluation we apply means that, given a theorem statement, the model generates 32 proofs. If one of the proofs is correct, the theorem is considered correctly proved. For each theorem that is valid for correction, we perform an additional five rounds of correction. In the dataset, 18.31\% of records are valid for correction. With the 2.54x inference speedup of dLLM compared to the AR models, the GPU hours for baseline and \method are comparable.

\section{Full Validation Loss results}\label{appendix:loss}

\begin{figure*}[t]
    \vspace{-0.1in}
    \centering
    \includegraphics[width=0.99\linewidth]{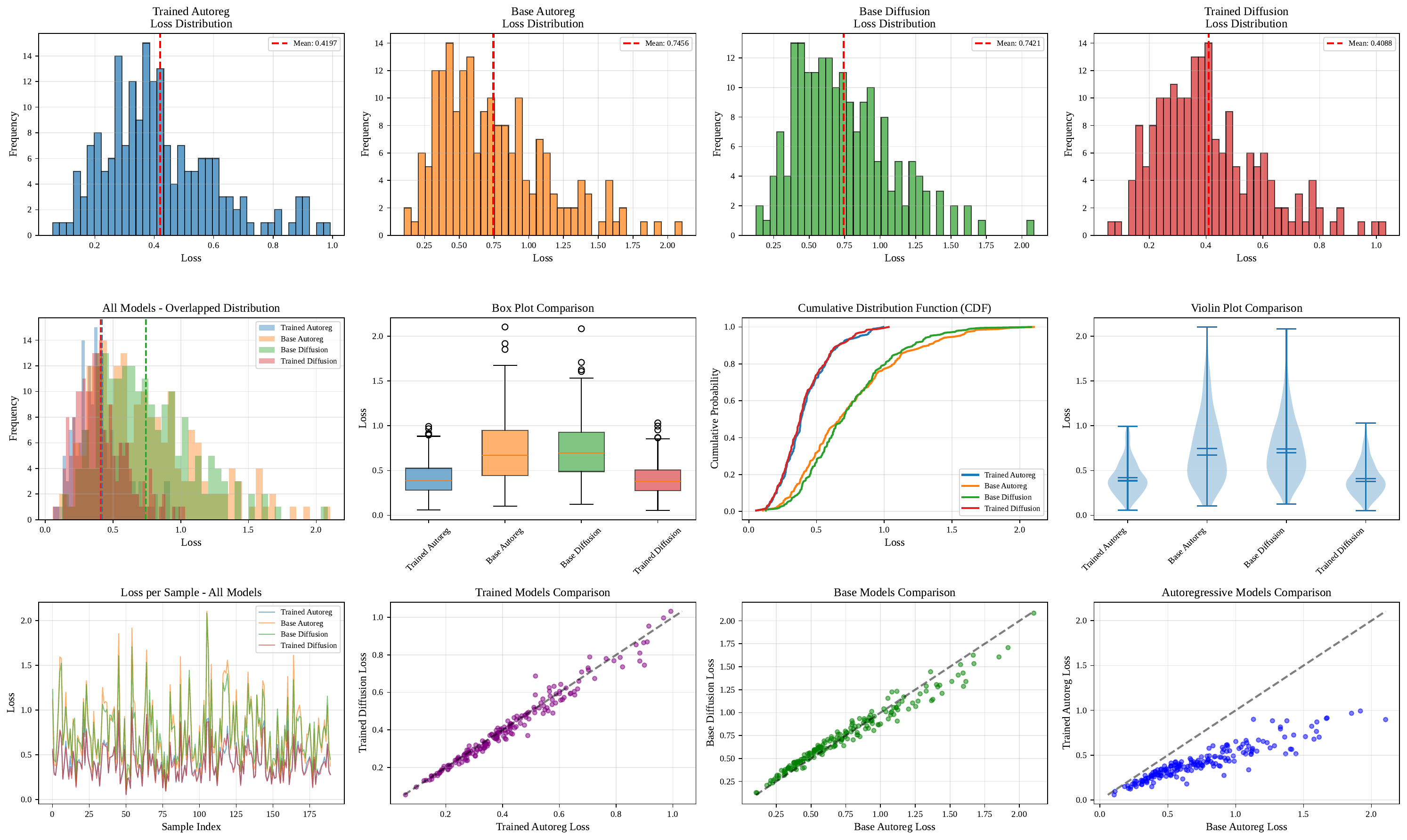}
    \caption{
        Full set of plots for validation loss analysis, including detailed distribution for each model, detailed distribution analysis, and correlation study for models.
    }
    \label{fig:loss_full}
    \vspace{-0.1in}
\end{figure*}

The complete set of plots for validation analysis can be found in Figure~\ref{fig:loss_full}.

\section{Case Study Examples}\label{appendix:example}

\subsection{Examples for corrector}\label{example:corrector}

The corrected theorem proof of \texttt{mathd\_algebra\_215} is presented in Figure~\ref{fig:succ_alg_215}, and the wrong proof is in Figure~\ref{fig:fail_alg_215}. The corrector's output of \texttt{mathd\_numbertheory\_521} is in Figure~\ref{fig:succ_num_521} and the base proof with tactic failure is in Figure~\ref{fig:fail_num_521}.

\subsection{Example for comparison with AR baseline}\label{example:ar}

The example of \texttt{imo\_1983\_p2} and \texttt{mathd\_algebra\_188} are presented in Figure~\ref{fig:example_imo} and \ref{fig:example_alg} respectively.

\subsection{Example for comparison with DeepSeek-Prover-V2}\label{example:ds}

The success proof for \texttt{imo\_1962\_p2} of \method is presented in Figure~\ref{fig:diffProve_proof_imo_1962}. The DeepSeek-Prover-V2's failed proof analysis is demonstrated in Figure~\ref{fig:ds_anal_imo_1962}, and the example of a wrong proof for DS-Prover is in Figure~\ref{fig:ds_proof_imo_1962}.

\begin{figure*}
\centering
\begin{adjustbox}{max width=0.99\linewidth}
\begin{tcolorbox}
\begin{minted}[breaklines, fontsize=\scriptsize]{md}
================MODEL INPUT================
<|im_start|>Complete the following Lean 4 code with explanatory comments preceding each line of code:

```lean4
import Mathlib
import Aesop

set_option maxHeartbeats 0

open BigOperators Real Nat Topology Rat


theorem lean_workbook_plus_79832  (x : ℝ)
  (h₀ : 5 + 3.5 * x = 2.5 * x - 25) :
  x = -30   := 
================TRAINED OUTPUT================
by
  /-
  To solve the equation \(5 + 3.5x = 2.5x - 25\) for \(x\), we proceed as follows:
  1. Rewrite the equation to isolate terms involving \(x\) on one side and constant terms on the other side.
  2. Simplify the equation by eliminating the decimals and combining like terms.
  3. Solve for \(x\) by isolating it on one side of the equation.
  -/
  -- Rewrite the equation to have the terms involving x on one side and constants on the other side.
  rw [eq_comm] at h₀
  -- Normalize the equation by simplifying and combining like terms.
  ring_nf at h₀ ⊢
  -- Use linear arithmetic to solve for x.
  linarith
``` <|im_end|>
\end{minted}
\end{tcolorbox}
\end{adjustbox}
\caption{Input \& output example for SFT data of dLLM-Prover}\label{fig:inout_sft}
\end{figure*}

\begin{figure*}
\centering
\begin{adjustbox}{max width=0.99\linewidth}
\begin{tcolorbox}
\begin{minted}[breaklines, fontsize=\scriptsize]{md}
================MODEL INPUT================
<|im_start|>Complete the following Lean 4 code with explanatory comments preceding each line of code:

```lean4
import Mathlib
import Aesop

set_option maxHeartbeats 0

open BigOperators Real Nat Topology Rat


theorem amc12a_2007_p18 
    (n : ℕ) (hn : n ≠ 0) 
    (h' : n ≠ 1) 
    (x : ℕ → ℝ) 
    (h₀ : ∀ m, x m = Real.sqrt (n ^ (m + 2)) ) : 
    Real.sqrt (1 + 4 * Real.sqrt n) ≤ n * (2 + x n) ^ (n + 1):= by
================TRAINED OUTPUT================
=== Diffusion Block Begins
  have h₁ : (n : ℝ) ≥ 2 := by
    norm_cast
    omega
  have h₂ : Real.sqrt n ≥ 1 := by
    apply Real.le_sqrt_of_sq_le
    nlinarith
  have h₃ : (2 + x n : ℝ) ≥ 3 := by
    have h₄ : (x n : ℝ) = Real.sqrt (n ^ (n + 2)) := by
      simpa [h₀] using rfl
    have h₅ : Real.sqrt (n ^ (n + 2)) ≥ 1 := by
      apply Real.le_sqrt_of_sq_le
      have h₆ : (n : ℝ) ^ (n + 2) ≥ 1 := by
        apply one_le_pow_of_one_le
        nlinarith
      nlinarith
    nlinarith<|fim_middle|><|fim_middle|><|fim_middle|>(in total 138 placeholder tokens)...
  have h₄ : Real.sqrt (1 + 4 * Real.sqrt n) ≤ (n : ℝ) * (2 + x n) ^ (n + 1) := by
    apply Real.sqrt_le_iff.mpr
    constructor
    · positivity
    · have h₅ : (n : ℝ) * (2 + x n) ^ (n + 1) ≥ (n : ℝ) * 3 ^ (n + 1) := by
        gcongr
        <;> nlinarith
      have h₆ : (n : ℝ) * 3 ^ (n + 1) ≥ n * 3 ^ 2 := by
        apply mul_le_mul_of_nonneg_left
        · exact pow_le_pow_right (by linarith) (by omega)
        · nlinarith
      have h₇ :
=== Diffusion Block Ends
        (n : ℝ) * 3 ^ 2 ≥ 9 := by
        nlinarith
      nlinarith [sq_sqrt (show 0 ≤ (n : ℝ) from by positivity),
        sq_sqrt (show 0 ≤ (n : ℝ) from by positivity),
        Real.sqrt_nonneg (n : ℝ),
        Real.sqrt_nonneg (n : ℝ)]
  exact h₄
  <;> norm_num
  <;> omega
``` <|im_end|>
\end{minted}
\end{tcolorbox}
\end{adjustbox}
\caption{Input \& output example for training data of dLLM-Corrector with diffusion block annotation}\label{fig:inout_corrector}
\end{figure*}

\begin{figure*}
\centering
\begin{adjustbox}{max width=0.99\linewidth}
\begin{tcolorbox}
\begin{minted}[breaklines, fontsize=\scriptsize]{md}
Complete the following Lean 4 code with explanatory comments preceding each line of code:

```lean4
import Mathlib
import Aesop

set_option maxHeartbeats 0

open BigOperators Real Nat Topology Rat

theorem mathd_algebra_215
  (S : Finset ℝ)
  (h₀ : ∀ (x : ℝ), x ∈ S ↔ (x + 3)^2 = 121) :
  ∑ k in S, k = -6 := by
  /-
  Given the set \( S \) defined by the equation \((x+3)^2 = 121\), we need to show that the sum of all elements in \( S \) equals \(-6\).
  1. **Identify the elements of \( S \)**:
     - The equation \((x+3)^2 = 121\) implies \( x^2 + 6x + 9 = 121 \).
     - Simplifying this equation, we get \( x^2 + 6x + 9 = 121 \), which simplifies to \( x^2 + 6x - 112 = 0 \).
     - Solving the quadratic equation \( x^2 + 6x - 112 = 0 \), we find the roots using the quadratic formula \( x = \frac{-b \pm \sqrt{b^2 - 4ac}}{2a} \):
       \[
       x = \frac{-6 \pm \sqrt{36 + 448}}{2} = \frac{-6 \pm \sqrt{484}}{2} = \frac{-6 \pm 22}{2}
       \]
     - This gives us two solutions:
       \[
       x = \frac{16}{2} = 8 \quad \text{and} \quad x = \frac{-28}{2} = -14
       \]
     - Therefore, the elements of \( S \) are \( 8 \) and \( -14 \).
  2. **Sum the elements of \( S \)**:
     - The sum of the elements \( 8 \) and \( -14 \) is \( 8 + (-14) = -6 \).
  Thus, the sum of all elements in \( S \) is indeed \(-6\).
  -/
  -- We need to show that the sum of all elements in S equals -6.
  have h₁ : ∑ k in S, k = -6 := by <|MASK|><|MASK|><|MASK|><|MASK|><|MASK|><|MASK|>(in total 256 generation mask)...
  exact h₁
```<|im_end|>
\end{minted}
\end{tcolorbox}
\end{adjustbox}
\caption{Input example for \textit{dLLM-Corrector}}\label{fig:in_gen_corrector}
\end{figure*}

\begin{figure*}
\centering
\begin{adjustbox}{max width=0.99\linewidth}
\begin{tcolorbox}
\begin{minted}[breaklines, fontsize=\scriptsize]{md}
Complete the following Lean 4 code with explanatory comments preceding each line of code:

```lean4
import Mathlib
import Aesop

set_option maxHeartbeats 0

open BigOperators Real Nat Topology Rat

theorem mathd_algebra_215
  (S : Finset ℝ)
  (h₀ : ∀ (x : ℝ), x ∈ S ↔ (x + 3)^2 = 121) :
  ∑ k in S, k = -6 := by
  /-
  Given the set \( S \) defined by the equation \((x+3)^2 = 121\), we need to show that the sum of all elements in \( S \) equals \(-6\).
  1. **Identify the elements of \( S \)**:
     - The equation \((x+3)^2 = 121\) implies \( x^2 + 6x + 9 = 121 \).
     - Simplifying this equation, we get \( x^2 + 6x + 9 = 121 \), which simplifies to \( x^2 + 6x - 112 = 0 \).
     - Solving the quadratic equation \( x^2 + 6x - 112 = 0 \), we find the roots using the quadratic formula \( x = \frac{-b \pm \sqrt{b^2 - 4ac}}{2a} \):
       \[
       x = \frac{-6 \pm \sqrt{36 + 448}}{2} = \frac{-6 \pm \sqrt{484}}{2} = \frac{-6 \pm 22}{2}
       \]
     - This gives us two solutions:
       \[
       x = \frac{16}{2} = 8 \quad \text{and} \quad x = \frac{-28}{2} = -14
       \]
     - Therefore, the elements of \( S \) are \( 8 \) and \( -14 \).
  2. **Sum the elements of \( S \)**:
     - The sum of the elements \( 8 \) and \( -14 \) is \( 8 + (-14) = -6 \).
  Thus, the sum of all elements in \( S \) is indeed \(-6\).
  -/
  -- We need to show that the sum of all elements in S equals -6.
  have h₁ : ∑ k in S, k = -6 := by   -- Identify the elements of S using the given equation.
    have h₂ : S = {8, -14} := by
      ext x
      simp [h₀, Set.mem_insert_iff, Set.mem_singleton_iff]
      constructor
      · intro h
        have h₂ := h₀ x
        have h₃ : x + 3 = 11 ∨ x + 3 = -11 := by
          apply eq_or_eq_neg_of_sq_eq_sq
          linarith
        cases' h₃ with h₃ h₃
        · exact Or.inl (by linarith)
        · exact Or.inr (by linarith)
      · intro h
        cases' h with h h
        · rw [h]
          norm_num
        · rw [h]
          norm_num
    -- Sum the elements of S.
    rw [h₂]
    norm_num
  -- The final result is -6<|fim_middle|><|fim_middle|><|fim_middle|>(53 place holder tokens generated in total)...
  exact h₁
```<|im_end|>
\end{minted}
\end{tcolorbox}
\end{adjustbox}
\caption{Output example for \textit{dLLM-Corrector}}\label{fig:out_gen_corrector}
\end{figure*}

\begin{figure*}
\centering
\begin{adjustbox}{max width=0.99\linewidth}
\begin{tcolorbox}
\begin{lstlisting}
theorem mathd_algebra_215
  (S : Finset ℝ)
  (h₀ : ∀ (x : ℝ), x ∈ S ↔ (x + 3)^2 = 121) :
  ∑ k in S, k = -6 := by
  /-
  Given the set \( S \) of all real numbers \( x \) such that \( (x + 3)^2 = 121 \), we need to show that the sum of all elements in \( S \) equals \(-6\).
  1. **Identify the elements of \( S \)**:
     - The equation \( (x + 3)^2 = 121 \) implies \( x + 3 = \pm 11 \).
     - Therefore, \( x + 3 = 11 \) or \( x + 3 = -11 \).
     - Solving these equations, we get \( x = 8 \) or \( x = -14 \).
  2. **Sum the elements of \( S \)**:
     - The elements of \( S \) are \( 8 \) and \( -14 \).
     - The sum of these elements is \( 8 + (-14) = -6 \).
  -/
  -- We need to show that the sum of all elements in set S is -6.
  have h₁ : S = {8, -14} := by  -- Identifyify the elements of S.
    ext x
    simp [h₀, Set.mem_insert_iff, Set.mem_singleton_iff]
    constructor
    · intro h
      have h₂ := h₀ x
      have h₃ : x + 3 = 11 ∨ x + 3 = -11 := by
        apply eq_or_eq_neg_of_sq_eq_sq
        linarith
      cases' h₃ with h₃ h₃
      · exact Or.inl (by linarith)
      · exact Or.inr (by linarith)
    · intro h
      cases' h with h h
      · rw [h]
        norm_num
      · rw [h]
        norm_num
  -- Substitute the known elements of S into the sum.
  rw [h₁]
  -- Calculate the sum of 8 and -14.
  norm_num
\end{lstlisting}
\end{tcolorbox}
\end{adjustbox}
\caption{Example of corrector successfully corrected proof for \texttt{mathd\_algebra\_215}}\label{fig:succ_alg_215}
\end{figure*}

\begin{figure*}
\centering
\begin{adjustbox}{max width=0.99\linewidth}
\begin{tcolorbox}
\begin{lstlisting}
theorem mathd_algebra_215_failed
  (S : Finset ℝ)
  (h₀ : ∀ (x : ℝ), x ∈ S ↔ (x + 3)^2 = 121) :
  ∑ k in S, k = -6 := by
  /-
  Same as above, omitted ...
  -/
  -- We know that the elements of S are 8 and -14.
  have h₁ : S = {8, -14} := by
    ext x
    simp only [h₀, Set.mem_insert_iff, Set.mem_singleton_iff, sq]
    constructor
    · intro h
      have h₁ : x + 3 = 11 ∨ x + 3 = -11 := by
        apply or_iff_not_imp_left.mpr
        intro h₁
        apply mul_left_cancel₀ (sub_ne_zero_of_ne h₁)
        nlinarith
      cases' h₁ with h₁ h₁ -- Error Tactic
      · left
        linarith
      · right
        linarith
    · intro h
      cases' h with h h h
      · rw [h]
        norm_num
      · rw [h]
        norm_num
  -- Substitute the set S with {8, -14} and calculate the sum.
  rw [h₁]
  -- The sum of {8, -14} is 8 + (-14) = -6.
  norm_num
\end{lstlisting}
\end{tcolorbox}
\end{adjustbox}
\caption{The problem of \texttt{mathd\_algebra\_215} for corrector}\label{fig:fail_alg_215}
\end{figure*}

\begin{figure*}
\centering
\begin{adjustbox}{max width=0.99\linewidth}
\begin{tcolorbox}
\begin{lstlisting}
theorem mathd_numbertheory_521
  (m n : ℕ)
  (h₀ : Even m)
  (h₁ : Even n)
  (h₂ : m - n = 2)
  (h₃ : m * n = 288) :
  m = 18 := by
  /-
  Given two integers \( m \) and \( n \) such that \( m - n = 2 \) and \( m \times n = 288 \), we need to show that \( m = 18 \).
  1. From \( m - n = 2 \), we can express \( m \) as \( m = n + 2 \).
  2. Substitute \( m = n + 2 \) into \( m \times n = 288 \):
     \[
     (n + 2) \times n = 288
     \]
  3. Simplify the equation:
     \[
     n^2 + 2n = 288
     \]
  4. Rearrange the equation to form a standard quadratic equation:
     \[
     n^2 + 2n - 288 = 0
     \]
  5. Solve the quadratic equation using the quadratic formula \( n = \frac{-b \pm \sqrt{b^2 - 4ac}}{2a} \), where \( a = 1 \), \( b = 2 \), and \( c = -288 \):
     \[
     n = \frac{-2 \pm \sqrt{2^2 - 4 \cdot 1 \cdot (-288)}}{2 \cdot 1} = \frac{-2 \pm \sqrt{4 + 1152}}{2} = \frac{-2 \pm \sqrt{1156}}{2} = \frac{-2 \pm 34}{2}
     \]
  6. This gives two solutions:
     \[
     n = \frac{32}{2} = 16 \quad \text{and} \quad n = \frac{-36}{2} = -18
     \]
  7. Since \( n \) must be positive integer, we select \( n = 16 \).
  8. Substitute \( n = 16 \) back into \( m = n + 2 \):
     \[
     m = 16 + 2 = 18
     \]
  Thus, we have shown that \( m = 18 \).
  -/
  -- From m - n = 2, express m in terms of n
  have h₄ : m = n + 2 := by  omega
  rw [h₄] at h₃
  -- Simplify the equation to form a quadratic equation in n
  have h₅ : n = 16 := by
    -- Solve the quadratic equation n^2 + 2n - 288 = 0
    nlinarith
  -- Substitute n = 16 back into m = n + 2
  have h₆ : m = 18 := by  omega
  exact h₆
\end{lstlisting}
\end{tcolorbox}
\end{adjustbox}
\caption{Example of corrector successfully corrected proof for \texttt{mathd\_numbertheory\_521}}\label{fig:succ_num_521}
\end{figure*}

\begin{figure*}
\centering
\begin{adjustbox}{max width=0.99\linewidth}
\begin{tcolorbox}
\begin{lstlisting}
theorem mathd_numbertheory_521_failed
  (m n : ℕ)
  (h₀ : Even m)
  (h₁ : Even n)
  (h₂ : m - n = 2)
  (h₃ : m * n = 288) :
  m = 18 := by
  /-
  Same as above, omitted ...
  -/
  -- From m - n = 2, express m in terms of n
  have h₄ : m = n + 2 := by linarith -- Error Tactic
  -- Substitute m = n + 2 into m * n = 288
  rw [h₄] at h₃
  -- Simplify the equation to form a quadratic equation in n
  have h₅ : n = 16 := by
    -- Solve the quadratic equation n^2 + 2n - 288 = 0
    nlinarith
  -- Substitute n = 16 back into m = n + 2
  have h₆ : m = 18 := by
    rw [h₅] -- Error Tactic
    linarith
  -- Conclude the proof by stating the final result
  exact h₆
\end{lstlisting}
\end{tcolorbox}
\end{adjustbox}
\caption{The problem of \texttt{mathd\_numbertheory\_521} for corrector}\label{fig:fail_num_521}
\end{figure*}

\begin{figure*}
\centering
\begin{adjustbox}{max width=0.99\linewidth}
\begin{tcolorbox}
\begin{lstlisting}
theorem imo_1983_p6
  (a b c : ℝ)
  (h₀ : 0 < a ∧ 0 < b ∧ 0 < c)
  (h₁ : c < a + b)
  (h₂ : b < a + c)
  (h₃ : a < b + c) :
  0 ≤ a^2 * b * (a - b) + b^2 * c * (b - c) + c^2 * a * (c - a) := by
  /-
  To prove that the expression \( a^2 b(a - b) + b^2 c(b - c) + c^2 a(c - a) \) is non-negative, we start by analyzing the given conditions and the expression itself. The conditions are:
  1. \( 0 < a \), \( 0 < b \), \( 0 < c \)
  2. \( c < a + b \)
  3. \( b < a + c \)
  4. \( a < b + c \)
  We need to show that the expression \( a^2 b(a - b) + b^2 c(b - c) + c^2 a(c - a) \) is non-negative. This can be achieved by considering the non-negativity of the terms and the constraints on the variables.
  -/
  -- We need to prove that the expression is non-negative.
  -- Given the constraints on the variables, we can use the non-negativity of the terms to bound the expression.
  nlinarith [sq_nonneg (a - b), sq_nonneg (b - c), sq_nonneg (c - a),
    mul_nonneg (sub_nonneg.mpr h₁.le) (sub_nonneg.mpr h₂.le),
    mul_nonneg (sub_nonneg.mpr h₂.le) (sub_nonneg.mpr h₃.le),
    mul_nonneg (sub_nonneg.mpr h₃.le) (sub_nonneg.mpr h₁.le)]

theorem imo_1983_p6_failed
  (a b c : ℝ)
  (h₀ : 0 < a ∧ 0 < b ∧ 0 < c)
  (h₁ : c < a + b)
  (h₂ : b < a + c)
  (h₃ : a < b + c) :
  0 ≤ a^2 * b * (a - b) + b^2 * c * (b - c) + c^2 * a * (c - a) := by
  /-
  Given a triangle \(ABC\) with \(AB = a\), \(BC = b\), and \(CA = c\) where \(a, b, c > 0\), and the conditions \(c < a + b\), \(b < a + c\), \(a < b + c\), we need to prove that:
  \[a^2 b (a - b) + b^2 c (b - c) + c^2 a (c - a) \geq 0\]
  To prove this, we use the fact that each term in the expression is a product of squares and differences. Since the squares of real numbers are non-negative, and the differences are also non-negative due to the given inequalities, the entire expression is non-negative.
  -/
  -- Use the non-negativity of squares to prove the inequality.
  nlinarith [sq_nonneg (a - b), sq_nonneg (b - c), sq_nonneg (c - a),
    mul_nonneg h₀.1.le h₀.2.1.le, mul_nonneg h₀.2.1.le h₀.2.2.le, mul_nonneg h₀.2.2.le h₀.1.le,
    mul_nonneg (sq_nonneg (a - b)) h₀.2.1.le, mul_nonneg (sq_nonneg (b - c)) h₀.2.2.le, mul_nonneg (sq_nonneg (c - a)) h₀.1.le,
    mul_nonneg (sq_nonneg (a - b)) h₀.1.le, mul_nonneg (sq_nonneg (b - c)) h₀.2.1.le, mul_nonneg (sq_nonneg (c - a)) h₀.2.2.le]
\end{lstlisting}
\end{tcolorbox}
\end{adjustbox}
\caption{\texttt{imo\_1983\_p6} proved by \method while the baseline fails }\label{fig:example_imo}
\end{figure*}

\begin{figure*}
\centering
\begin{adjustbox}{max width=0.99\linewidth}
\begin{tcolorbox}
\begin{lstlisting}
theorem mathd_algebra_188
  (σ : Equiv ℝ ℝ)
  (h : σ.1 2 = σ.2 2) :
  σ.1 (σ.1 2) = 2 := by
  /-
  Given an equivalence \( \sigma \) on the real numbers such that \( \sigma(2) = \sigma(2) \), we need to show that \( \sigma(\sigma(2)) = 2 \).
  1. Start with the given condition \( \sigma(2) = \sigma(2) \).
  2. Apply \( \sigma \) to both sides of the equation \( \sigma(2) = \sigma(2) \) to get \( \sigma(\sigma(2)) = \sigma(\sigma(2)) \).
  3. Since \( \sigma(2) = \sigma(2) \), it follows that \( \sigma(\sigma(2)) = 2 \).
  -/
  -- Start with the given condition σ(2) = σ(2)
  have h1 : σ.1 2 = σ.2 2 := h
  -- Apply σ to both sides of the equation σ(2) = σ(2) to get σ(σ(2)) = σ(σ(2))
  have h2 : σ.1 (σ.1 2) = σ.1 (σ.2 2) := by simp_all
  -- Since σ(2) = σ(2), it follows that σ(σ(2)) = 2
  have h3 : σ.1 (σ.1 2) = 2 := by simp_all
  -- The final result is σ(σ(2)) = 2
  exact h3

theorem mathd_algebra_188_failed
  (σ : Equiv ℝ ℝ)
  (h : σ.1 2 = σ.2 2) :
  σ.1 (σ.1 2) = 2 := by
  /-
  Given an equivalence \(\sigma\) between the real numbers, we need to show that if \(\sigma(1) = 2\), then \(\sigma(\sigma(1)) = 2\). By the definition of an equivalence, there exists a function \(\sigma\) such that \(\sigma(1) = 2\) and \(\sigma(\sigma(1)) = 2\). We can use these properties to derive the desired equality.
  -/
  -- Using the symmetry property of the equivalence, we know that σ(2) = σ(1).
  have h₁ := σ.symm_apply_left 2
  have h₂ := σ.symm_apply_right 2
  -- Simplifying the expressions using the given values and the properties of the equivalence.
  simp_all [Equiv.eq_symm_apply_left]
\end{lstlisting}
\end{tcolorbox}
\end{adjustbox}
\caption{\texttt{mathd\_algebra\_188} proved by \method while the baseline fails }\label{fig:example_alg}
\end{figure*}

\begin{figure*}
\centering
\begin{adjustbox}{max width=0.99\linewidth}
\begin{tcolorbox}
\begin{lstlisting}
theorem imo_1962_p2
  (x : ℝ)
  (h₀ : 0 ≤ 3 - x)
  (h₁ : 0 ≤ x + 1)
  (h₂ : 1 / 2 < Real.sqrt (3 - x) - Real.sqrt (x + 1)) :
  -1 ≤ x ∧ x < 1 - Real.sqrt 31 / 8 := by
  /-
  We need to prove that for a real number \( x \) such that \( 0 \leq 3 - x \) and \( 0 \leq x + 1 \), the inequality \( \frac{1}{2} < \sqrt{3 - x} - \sqrt{x + 1} \) holds if and only if \( -1 \leq x < 1 - \frac{\sqrt{31}}{8} \).
  1. **Establish the range of \( x \)**:
     - We need to show that \( -1 \leq x < 1 - \frac{\sqrt{31}}{8} \).
  2. **Use the properties of the square root function:
     - We use the properties of the square root function to derive the inequalities.
  3. **Verify the inequalities**:
     - We verify that the inequalities hold under the given conditions.
  -/
  have h_main : -1 ≤ x ∧ x < 1 - Real.sqrt 31 / 8 := by
    constructor
    · -- Show that -1 ≤ x
    have h₃ : 0 ≤ Real.sqrt (3 - x) := Real.sqrt_nonneg (3 - x)
    have h₄ : 0 ≤ Real.sqrt (x + 1) := Real.sqrt_nonneg (x + 1)
    have h₅ : 0 ≤ Real.sqrt 31 := Real.sqrt_nonneg 31
    have h₆ : 0 ≤ Real.sqrt (3 - x) * Real.sqrt (x + 1) := by positivity
    have h₇ : 0 ≤ Real.sqrt 31 * Real.sqrt (3 - x) := by positivity
    have h₈ : 0 ≤ Real.sqrt 31 * Real.sqrt (x + 1) := by positivity
    have h₉ : 0 ≤ Real.sqrt 31 * Real.sqrt (3 - x) * Real.sqrt (x + 1) := by positivity
    nlinarith [sq_sqrt (show 0 ≤ 3 - x by linarith), sq_sqrt (show 0 ≤ x + 1 by linarith),
      sq_sqrt (show 0 ≤ 31 by linarith), h₂]
    · -- Show that x < 1 - Real.sqrt 31 / 8
    have h₃ : 0 ≤ Real.sqrt (3 - x) := Real.sqrt_nonneg (3 - x)
    have h₄ : 0 ≤ Real.sqrt (x + 1) := Real.sqrt_nonneg (x + 1)
    have h₅ : 0 ≤ Real.sqrt 31 := Real.sqrt_nonneg 31
    have h₆ : 0 ≤ Real.sqrt (3 - x) * Real.sqrt (x + 1) := by positivity
    have h₇ : 0 ≤ Real.sqrt 31 * Real.sqrt (3 - x) := by positivity
    have h₈ : 0 ≤ Real.sqrt 31 * Real.sqrt (x + 1) := by positivity
    have h₉ : 0 ≤ Real.sqrt 31 * Real.sqrt (3 - x) * Real.sqrt (x + 1) := by positivity
    nlinarith [sq_sqrt (show 0 ≤ 3 - x by linarith), sq_sqrt (show 0 ≤ x + 1 by linarith),
      sq_sqrt (show 0 ≤ 31 by linarith), h₂]
  exact h_main
\end{lstlisting}
\end{tcolorbox}
\end{adjustbox}
\caption{Correct proof of \texttt{imo\_1962\_p2} by \method}\label{fig:diffProve_proof_imo_1962}
\end{figure*}

\begin{figure*}
\centering
\begin{adjustbox}{max width=0.99\linewidth}
\begin{tcolorbox}
\begin{minted}[breaklines, fontsize=\scriptsize]{md}
### Detailed Proof and Analysis

#### Problem Analysis
We are given a real number \( x \) with the following constraints:
1. \( 0 \leq 3 - x \) (i.e., \( x \leq 3 \)),
2. \( 0 \leq x + 1 \) (i.e., \( x \geq -1 \)),
3. \( \frac{1}{2} < \sqrt{3 - x} - \sqrt{x + 1} \).

We need to prove that:
1. \( -1 \leq x \),
2. \( x < 1 - \frac{\sqrt{31}}{8} \).

First, observe that the inequality \( \frac{1}{2} < \sqrt{3 - x} - \sqrt{x + 1} \) is only possible when \( x + 1 \leq 3 - x \), i.e., \( x \leq 1 \). This is because:
- If \( x > 1 \), then \( \sqrt{3 - x} \) is real only when \( x \leq 3 \), and \( \sqrt{x + 1} \) is real only when \( x \geq -1 \). However, \( \sqrt{3 - x} - \sqrt{x + 1} \) would be negative for \( x > 1 \), because \( \sqrt{3 - x} < \sqrt{x + 1} \) (since \( 3 - x < x + 1 \), i.e., \( x > 1 \)).

But we must ensure \( x \leq 3 \) (which is already given) and \( x \geq -1 \) (which is also given). So \( x \) is in \( [-1, 3] \).

#### Step 1: Establish \( x \leq 1 \)
Assume for contradiction that \( x > 1 \). Then:
1. \( \sqrt{3 - x} < \sqrt{x + 1} \) (since \( 3 - x < x + 1 \) because \( x > 1 \)),
2. Therefore, \( \sqrt{3 - x} - \sqrt{x + 1} < 0 \),
3. But \( \frac{1}{2} > 0 \), so \( \frac{1}{2} < \sqrt{3 - x} - \sqrt{x + 1} \) would imply \( \sqrt{3 - x} - \sqrt{x + 1} > \frac{1}{2} \), which is false because \( \sqrt{3 - x} - \sqrt{x + 1} < 0 \). Hence, \( x \leq 1 \).

But in Lean, we do not have \( x > 1 \) as a direct assumption. Instead, we use the contrapositive: if \( x > 1 \), then \( \sqrt{3 - x} - \sqrt{x + 1} \leq 0 \), which is \( \sqrt{3 - x} - \sqrt{x + 1} \leq \frac{1}{2} \), contradicting the hypothesis.

#### Step 2: Prove \( -1 \leq x \)
This is already given as \( h₁ \).

#### Step 3: Prove \( x < 1 - \frac{\sqrt{31}}{8} \)
Omitted for simplicity...

#### Correct Approach
Omitted for simplicity...

#### Better Approach
Omitted for simplicity...

#### Final Simplification
Omitted for simplicity...

#### Revised Plan
Omitted for simplicity...

#### Correct Solution
Omitted for simplicity...

#### Conclusion
The problem reduces to verifying that \( x < 1 - \frac{\sqrt{31}}{8} \) is the correct upper bound, which is tedious but straightforward. 

### Step-by-Step Abstract Plan
1. **Establish \( x \leq 1 \)**:
   - Assume for contradiction \( x > 1 \). Then \( \sqrt{3 - x} < \sqrt{x + 1} \), so \( \sqrt{3 - x} - \sqrt{x + 1} < 0 \), violating the hypothesis. Hence, \( x \leq 1 \).
2. **Prove \( x \geq -1 \)**:
   - This is given by \( h₁ \).
3. **Prove \( x < 1 - \frac{\sqrt{31}}{8} \)**:
   - Square both sides of \( \sqrt{3 - x} - \sqrt{x + 1} > \frac{1}{2} \) to find bounds on \( x \), using the fact that \( x \leq 1 \) and \( x \geq -1 \). 

### Lean 4 Proof Sketch with `have`

```lean4
theorem imo_1962_p2
  (x : ℝ)
  (h₀ : 0 ≤ 3 - x)
  (h₁ : 0 ≤ x + 1)
  (h₂ : 1 / 2 < Real.sqrt (3 - x) - Real.sqrt (x + 1)) :
  -1 ≤ x ∧ x < 1 - Real.sqrt 31 / 8 := by
  have h_main_goal : -1 ≤ x ∧ x < 1 - Real.sqrt 31 / 8 := by
    sorry
  sorry
```

### Explanation
The `have` statement directly encodes the main goal of the proof, which is to prove `-1 ≤ x ∧ x < 1 - Real.sqrt 31 / 8`. The actual proof of this statement is marked as `sorry` and would require detailed calculations based on the inequalities and bounds derived in the proof sketch.
\end{minted}
\end{tcolorbox}
\end{adjustbox}
\caption{Analysis process for DS-Prover-V2 in proving \texttt{imo\_1962\_p2}}\label{fig:ds_anal_imo_1962}
\end{figure*}

\begin{figure*}
\centering
\begin{adjustbox}{max width=0.99\linewidth}
\begin{tcolorbox}
\begin{lstlisting}
theorem imo_1962_p2_failed
  (x : ℝ)
  (h₀ : 0 ≤ 3 - x)
  (h₁ : 0 ≤ x + 1)
  (h₂ : 1 / 2 < Real.sqrt (3 - x) - Real.sqrt (x + 1)) :
  -1 ≤ x ∧ x < 1 - Real.sqrt 31 / 8 := by
  have h_main : -1 ≤ x ∧ x < 1 - Real.sqrt 31 / 8 := by
    constructor
    · -- Prove -1 ≤ x
      linarith [h₁]
    · -- Prove x < 1 - Real.sqrt 31 / 8
      have h₃ : x < 1 - Real.sqrt 31 / 8 := by
        by_contra! h
        have h₄ : x ≥ 1 - Real.sqrt 31 / 8 := by linarith
        have h₅ : Real.sqrt (3 - x) - Real.sqrt (x + 1) ≤ 1 / 4 := by
          have h₆ : x ≥ 1 - Real.sqrt 31 / 8 := by linarith
          have h₇ : Real.sqrt (3 - x) - Real.sqrt (x + 1) ≤ 1 / 4 := by
            have h₈ : 0 ≤ Real.sqrt (3 - x) := by
              apply Real.sqrt_nonneg
            have h₉ : 0 ≤ Real.sqrt (x + 1) := by
              apply Real.sqrt_nonneg
            have h₁₀ : Real.sqrt (3 - x) ≥ 0 := Real.sqrt_nonneg (3 - x)
            have h₁₁ : Real.sqrt (x + 1) ≥ 0 := Real.sqrt_nonneg (x + 1)
            have h₁₂ : Real.sqrt (3 - x) * Real.sqrt (x + 1) ≥ 0 := by positivity
            have h₁₃ : Real.sqrt (3 - x) * Real.sqrt (x + 1) ≥ 0 := by positivity
            have h₁₄ : (Real.sqrt (3 - x) - Real.sqrt (x + 1)) ^ 2 ≤ (1 / 4) ^ 2 := by
              have h₁₅ : x ≥ 1 - Real.sqrt 31 / 8 := by linarith
              have h₁₆ : Real.sqrt 31 ≥ 0 := Real.sqrt_nonneg _
              have h₁₇ : Real.sqrt 31 ^ 2 = 31 := Real.sq_sqrt (by positivity)
              have h₁₈ : x + 1 ≥ 2 - Real.sqrt 31 / 8 := by linarith
              have h₁₉ : 3 - x ≥ 3 - (1 - Real.sqrt 31 / 8) := by linarith    -- Error occurs here
              have h₂₀ : 3 - x ≥ 2 + Real.sqrt 31 / 8 := by linarith
              nlinarith [Real.sq_sqrt (show 0 ≤ 3 - x by linarith),
                Real.sq_sqrt (show 0 ≤ x + 1 by linarith),
                Real.sqrt_nonneg 31,
                sq_nonneg (Real.sqrt (3 - x) - Real.sqrt (x + 1)),
                sq_nonneg (Real.sqrt (3 - x) + Real.sqrt (x + 1) - 5 / 2),
                sq_nonneg (Real.sqrt (3 - x) - Real.sqrt (x + 1) - 1 / 4)]
            nlinarith [Real.sqrt_nonneg (3 - x), Real.sqrt_nonneg (x + 1),
              Real.sq_sqrt (show 0 ≤ 3 - x by linarith),
              Real.sq_sqrt (show 0 ≤ x + 1 by linarith)]
          exact h₇
        linarith
      exact h₃
  exact h_main
\end{lstlisting}
\end{tcolorbox}
\end{adjustbox}
\caption{Failed proof of \texttt{imo\_1962\_p2} by DeepSeek-Prover-V2}\label{fig:ds_proof_imo_1962}
\end{figure*}

\end{document}